\documentclass[10pt,twocolumn,letterpaper]{article}

\usepackage{cvpr}
\usepackage{color}

\usepackage{graphicx}
\usepackage{amsmath}
\usepackage{amssymb}
\usepackage{color}

\usepackage[ruled,titlenotnumbered,linesnumbered,noline]{algorithm2e}

\newcommand{\IR}{\mathbb{R}}
\newcommand{\lengthS}{L(S)}
\newcommand{\dx}{\,\text{d}x}
\newcommand{\vecprod}[2]{\left\langle#1,#2\right\rangle}
\newcommand{\norm}[1]{\left\|#1\right\|}

\DeclareMathOperator{\Div}{div}

\newcommand{\exclude}[1]{}
\newcommand{\FS}[2]{%
	{\textcolor{red}{#1}}{\textcolor{blue}{#2}}%
	\-\marginpar[\raggedleft\footnotesize\bf{FS}]{\raggedright\footnotesize\bf{FS}}%
	}

 \makeatletter
\def\ifundefined{\@ifundefined}
\makeatother

\usepackage[pagebackref=true,breaklinks=true,letterpaper=true,colorlinks,bookmarks=false]{hyperref}

\usepackage[pagebackref=true,breaklinks=true,letterpaper=true,colorlinks,bookmarks=false]{hyperref}

\cvprfinalcopy 

\def\cvprPaperID{588} 

\ifcvprfinal\fi

\begin{document}

\title{An Experimental Comparison of {\em Trust Region} and {\em Level Sets}}


\author{
Lena Gorelick${}^{1}$
\and
Ismail Ben Ayed${}^{2,3}$
\and
Frank R. Schmidt${}^{4}$
\and
Yuri Boykov${}^{1}$
}

\maketitle

\setcounter{footnote}{0}
\footnotetext[1]{Computer Science Dept., Western University, Canada}
\footnotetext[2]{Medical Biophysics Dept., Western University, Canada}
\footnotetext[3]{GE Healthcare, Canada}
\footnotetext[4]{BIOSS Center of Biological Signalling Studies,}
{\let\thefootnote\relax\footnote{University of Freiburg, Germany}}
\setcounter{footnote}{0}

\begin{abstract}

High-order (non-linear) functionals have become very popular in
segmentation, stereo and other computer vision
problems. \emph{Level~sets} is a well established general gradient
descent framework, which is directly applicable to optimization of such
functionals and widely used in practice. Recently, another
general optimization approach based on {\em trust region} methodology
was proposed for regional non-linear functionals \cite{FTR:cvpr13}.
Our goal is a comprehensive experimental comparison of these two
frameworks in regard to practical efficiency, robustness to
parameters, and optimality. We experiment on a wide range of problems
with non-linear constraints on segment volume, appearance
and shape.

\end{abstract}

\section{Introduction}
    
We study a general class of complex non-linear segmentation energies
with high-order regional terms. Such energies are often desirable in
the computer vision tasks of image segmentation, co-segmentation and
stereo
\cite{kkz:iccv03,freedman:pami04,rother:cvpr06,Ismail:LevelSetsWithArea08,ismail:cvpr10,KC11:iccv,linesearchcuts:12,FTR:cvpr13}
and are particularly useful when there is a prior knowledge about
the appearance or the shape of an object being segmented.

We focus on  energies of the following form: 
\begin{equation} \label{eq:ERL}
\min_{S\in \Omega}E(S) = R(S) + \lambda \lengthS,
\end{equation}
where $S$ is a binary segmentation, $\lengthS$ is a standard
length-based smoothness term, and $R(S)$ is a (generally) non-linear
{\em regional functional} discussed below.

\begin{figure}[t]
\begin{center}
\includegraphics[width = 0.4\textwidth]{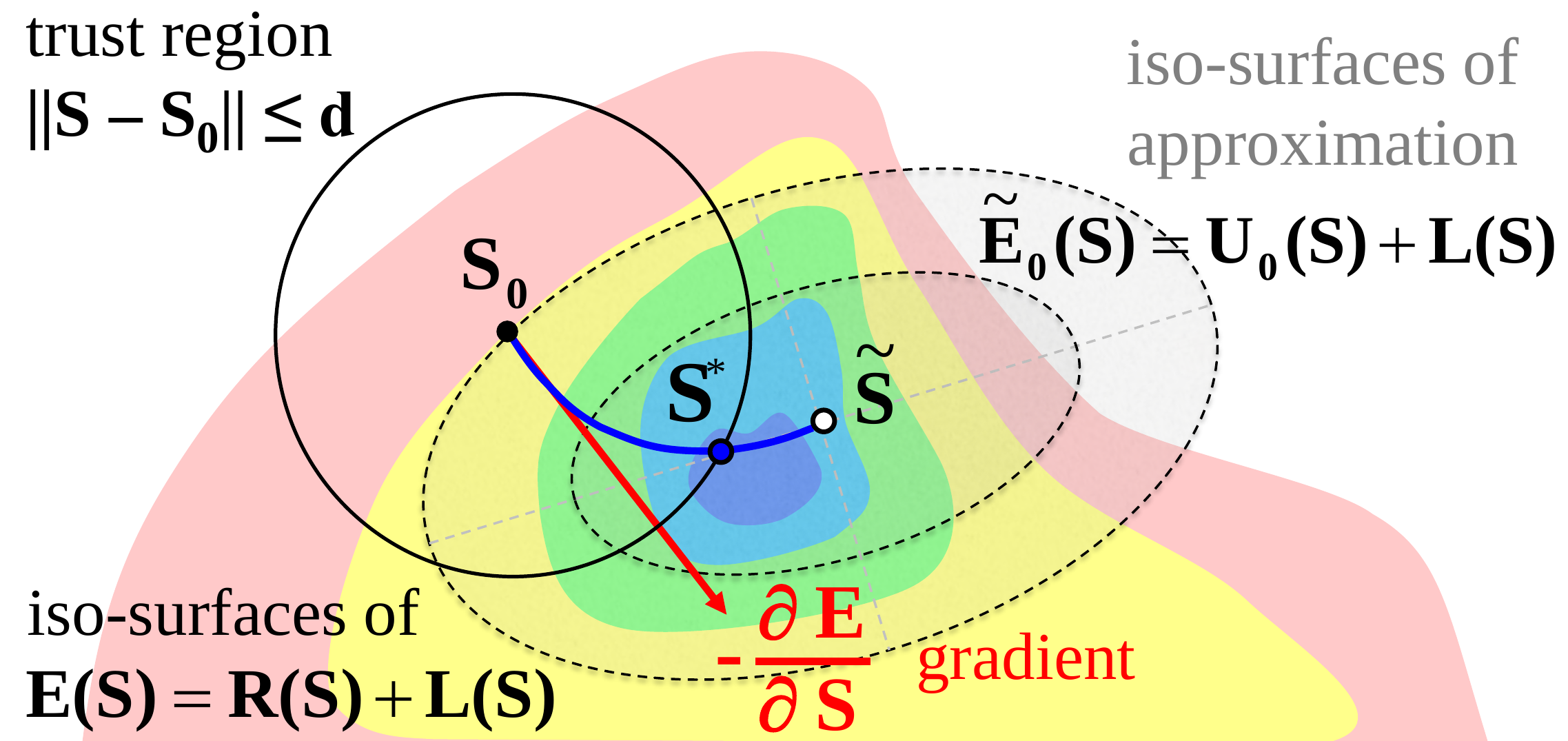}
\bf\caption{\rm%
  Gradient descent ({\em level sets}) and {\em trust region}. 
  \label{fig:yuri_lena}}
\end{center}
\end{figure}

Let $I:\Omega\to\IR^m$ be an image defined in $\Omega \subset \IR^n$.
The most basic type of regional terms used in segmentation is a linear
functional $U(S)$, which can be represented via an arbitrary scalar
function $f:\Omega\to\IR$
\begin{align}\label{eq:LinearRegionalFunctional}
U(S) = \int_S f\dx = \int_\Omega f\cdot 1_S\dx =: \vecprod{f}{S}.
\end{align}
Usually, $f$ corresponds to an {\em appearance model} based on image
intensities/colors $I$, e.g.,~$f(x)$ could be a log-likelihood
ratio for intensity $I(x)$ given particular object and background
intensity distributions. The integral in
\eqref{eq:LinearRegionalFunctional} can be seen as a dot product
between scalar function $f$ and $1_S$, which denotes the characteristic
function of set
$S$.  We use notation $\vecprod{f}{S}$ to refer to such linear
functionals.

More general \emph{non-linear regional functional} $R(S)$ can be
described as follows.  Assume $k$ scalar functions
$f_1,\ldots,f_k:\Omega\to\IR$, each defining a linear functional
$\vecprod{f_i}{S}$ of type~\eqref{eq:LinearRegionalFunctional}, and
one differentiable non-linear function
$F(v_1,\ldots,v_k)\colon\IR^k\to\IR$ that combines them,
\begin{align}\label{eq:NonlinearRegionalFunctional}
R(S) =     F\left(\vecprod{f_1}{S},\ldots,\vecprod{f_k}{S}\right).
\end{align}
Such general regional terms could enforce non-linear constraints on
volume or higher-order shape moments of segment $S$. They could also
penalize $L_2$ or other distance metrics between the distribution (or
non-normalized bin counts) of intensities/colors inside segment
$S$ and some given target. For example, $S$ could be softly
constrained to specific volume $V_0$ via quadratic functional
$$R(S) = (\vecprod{1}{S}-V_0)^2$$ using $f_1(x)=1$ and $F(v)=(v-V_0)^2$,
while the Kullback-Leibler (KL) divergence between the intensity distribution
in $S$ and a fixed target distribution $q=(q_1,\ldots,q_k)$ could be
written as
$$R(S) = \sum_i^k
\frac{\vecprod{f_i}{S}}{\vecprod{1}{S}}\log\left(\frac{\vecprod{f_i}{S}}{\vecprod{1}{S}\cdot
  q_i}\right).$$ Here scalar function $f_i(x)$ is an indicator for
pixels with intensity $i$ (in bin $i$).  More examples of non-linear
regional terms $R(S)$ are discussed in
Section~\ref{sec:experiments}\@.

In general, optimization of non-linear regional terms is NP-hard and
cannot be addressed by standard global optimization methods. Some
earlier papers developed specialized techniques for particular forms
of non-linear regional functionals.  For example, the algorithm in
\cite{ismail:cvpr10} was developed for minimizing the Bhattacharyya
distance between distributions and a dual-decomposition approach in
\cite{woodford:iccv09} applies to convex piece-wise linear regional
functionals \eqref{eq:NonlinearRegionalFunctional}.  These methods are
outside the scope of our paper since we focus on more general
techniques.  Combinatorial techniques \cite{kkz:iccv03,rother:cvpr06}
apply to general non-linear regional functionals, and they can make
large moves by globally minimizing some approximating energy.
However, as pointed out in \cite{linesearchcuts:12}, these methods are known to
converge to solutions that are not even a local minimum of energy
\eqref{eq:ERL}. Level sets are a well established in the litterature as a 
gradient descent framework that can address arbitrary differentiable functionals
\cite{LevelSetsBook:Ismail11}, and therefore, are widely used for
high-order terms
\cite{Adam2009,Foulonneau2009,BenAyed2009,Michailovich2007,freedman:pami04}.

This paper compares two known optimization methods applicable to
energy \eqref{eq:ERL} with any high-order regional term
\eqref{eq:NonlinearRegionalFunctional}: {\em trust region} and {\em
  level sets}. To address non-linear term $R(S)$, both methods use its
first-order functional derivative
\begin{align}\label{eq:dR}
\frac{\partial R}{\partial S} = 
    \sum_{i=1}^k   \frac{\partial F}{\partial v_i}
    \left(\vecprod{f_1}{S},\ldots,\vecprod{f_k}{S}\right) \cdot   f_i
\end{align}
either when computing gradient flow for \eqref{eq:ERL} in level sets
or when approximating energy \eqref{eq:ERL} in trust region. However,
despite using the same first derivative $\frac{\partial R}{\partial
  S}$, the level sets and trust region are fairly different approaches
to optimize~\eqref{eq:ERL}.

The structure of the paper is as follows.
Sections~\ref{sec:ls}-\ref{sec:tr} review the two general approaches that
we compare: gradient descent implemented with level sets and trust
region implemented with graph cuts. While the general trust region
framework \cite{FTR:cvpr13} can be based on a number of underlying
global optimization techniques, we specifically choose a graph cut
implementation (versus continuous convex relaxation approach), since
it is more appropriate for our CPU-based evaluations. Our experimental
results and comparisons are reported in Section~\ref{sec:experiments}
and Section~\ref{sec:conclusion} presents the conclusions.

\section{Overview of Algorithms}\label{sec:overview}
Below we provide general background information on level sets and
trust region methods, see Sections \ref{sec:ls}--\ref{sec:tr}.
High-level conceptual comparison of the two frameworks is provided in
Section \ref{sec:comp}.

\subsection{Standard Level Sets}\label{sec:ls}
In the level set framework, minimization of energy $E$ is carried out
by computing a partial differential equation (PDE), which governs the
evolution of the boundary of $S$. To this end, we derive the
Euler-Lagrange equation by embedding segment $S$ in a one-parameter
family $S(t)$, $t \in \mathbb{R}^{+}$, and solving a PDE of the
general form:
\begin{equation}
\label{eq:pde} \frac{\partial S}{\partial t} = -\frac{\partial E(S)}{\partial S} = - \frac{\partial R(S)}{\partial S} - \frac{\partial L(S)}{\partial S} 
\end{equation}
where $t$ is an artificial time step parameterizing the descent
direction. The basic idea is to describe segment $S$ implicitly via an
embedding function $\phi: \Omega \rightarrow \mathbb{R}$:
\begin{eqnarray}
\label{level_set_region_membership}
S &=& \{ x \in \Omega | \phi(x)\leq 0 \} \nonumber \\
\Omega \setminus S &=& \{ x \in \Omega | \phi(x) > 0 \},
\end{eqnarray} 
and evolve $\phi$ instead of $S$. With the above representation, the
terms that appear in energy (\ref{eq:ERL}) can be expressed as
functions of $\phi$ as follows \cite{Chan01ip,LevelSetsBook:Ismail11}:
\begin{eqnarray}\label{eq:LSlength}
L(S) &=& \int_{\Omega} \|\nabla H(\phi)\|dx = \int_{\Omega}\delta (\phi)\|\nabla \phi \|dx \nonumber \\
\vecprod{f_i}{S} &=& \int_\Omega H(\phi) f_i dx. 
\end{eqnarray}
Here, $\delta$ and $H$ denote the Dirac function and Heaviside
function, respectively.  Therefore, the evolution equation in
(\ref{eq:pde}) can be computed directly by applying the Euler-Lagrange
descent equation with respect to $\phi$. This gives the following
gradient flow:
\begin{equation}
\label{curve-flow-withoutSDF}
 \frac{\partial \phi}{\partial t} = 
  \left[-\frac{\partial R(S)}{\partial S}+\lambda\kappa\right]\delta(\phi)
\end{equation}
with $\kappa:=\Div\left(\frac{\nabla\phi}{\norm{\nabla\phi}}\right)$
denoting the curvature of $\phi$'s level lines. The first term in
(\ref{curve-flow-withoutSDF}) is a regional flow minimizing $R$, and
the second is a standard curvature flow minimizing the length of the
segment's boundary.

In standard level set implementations, it is numerically mandatory to
keep the evolving $\phi$ close to a distance function
\cite{Li2005a,Osher2002}. This can be done by re-initialization
procedures \cite{Osher2002}, which were intensively used in classical
level set methods \cite{Caselles97ijcv}. Such procedures, however,
rely on several {\em ad hoc} choices and may result in undesirable
side effects \cite{Li2005a}. In our implementation, we use an
efficient and well-known alternative \cite{Li2005a}, which adds an internal energy term that penalizes the deviation of $\phi$
from a distance function:
\begin{equation}
\label{distance-function-penalty}
\frac{\mu}{2} \int_{\Omega} \left (1 - \|\nabla \phi\| \right)^2 dx .
\end{equation}
In comparison to re-initialization procedures, the implementation in
\cite{Li2005a} allows larger time steps (and therefore faster curve
evolution).  Furthermore, it can be implemented via simple finite
difference schemes, unlike traditional level set implementations which
require complex upwind schemes \cite{Sethian1999}. With the
distance-function penalty, the gradient flow in
(\ref{curve-flow-withoutSDF}) becomes:
\begin{equation}
\label{curve-flow-withSDF}
 \frac{\partial \phi}{\partial t} = 
  \mu \left [ \Delta\phi - \kappa \right ] +
  \left[-\frac{\partial R(S)}{\partial S}+\lambda \kappa\right]\delta(\phi)  
\end{equation}
For all the experiments in this paper, we implemented the flow in
(\ref{curve-flow-withSDF}) using the numerical prescriptions in
\cite{Li2005a}.  For each point $p$ of the discrete grid, we update
the level set function as
\begin{equation}
\label{Discrete-Level-Set-Updates}
\phi^{j+1}(p) = \phi^{j}(p) + \Delta t\cdot A(\phi^{j}(p)),
\end{equation} 
where $\Delta t$ is the discrete time step and $j$ is the iteration
number.  $A(\phi^{j}(p))$ is a numerical approximation of the
right-hand side of (\ref{curve-flow-withSDF}), where the spatial
derivatives of $\phi$ are approximated with central differences and
the temporal derivative with forward differences.  The Dirac function
is approximated by $\delta_{\epsilon}(t) = \frac{1}{2\epsilon}
[1+\cos(\frac{\pi t}{\epsilon})]$ for $|t| \leq \epsilon$ and $0$
elsewhere. We use $\epsilon = 1.5$ and $\mu=0.05$.

  In the context of level set and PDE methods, it is known that the
  choice of time steps should follow strict numerical conditions to
  ensure stability of front propagation, e.g., the standard
  Courant-Friedrichs-Lewy (CFL) conditions \cite{Estellers2012}. These
  conditions require that $\Delta t$ should be smaller than a certain
  value $\tau$ that depends on the choice of discretization. The
  level set literature generally uses fixed time steps. For instance,
  classical upwind schemes \cite{Sethian1999} generally require a
  small $\Delta t$ for stability, whereas the scheme in \cite{Li2005a}
  allows relatively larger time steps. The optimum time step is not
  known a priori and finding a good $\Delta t < \tau$ via an adaptive
  search such as back-tracking \cite{Boyd2004} seems attractive. 
However,
 to
  apply a back-tracking scheme, we would have to evaluate the energy
  at each step. In the case of level sets, this requires a discrete
  approximation of the original continuous energy. We observed in our
  experiments that 
the gradient of such discrete 
	approximation of the energy does not
  coincide with the gradient obtained in the numerical updates in~
	\eqref{Discrete-Level-Set-Updates}.
  Therefore, with a back-tracking scheme, level sets get stuck very
  quickly in a local minimum of the discrete approximation of the
  energy 
(See the adaptive level set example in Fig. \ref{fig:volume}).
	We believe that this is the main reason why, to the best of
  our knowledge, back-tracking approaches are generally avoided in the
  level-set literature.  
Therefore, in the following level-set experiments, we use a standard scheme based on
fixed time step $\Delta t$ during each curve evolution, and report the performance at convergence
 for several values $\Delta t \in \{1 \ldots 10^3\}$.

\subsection{Trust Region Framework}\label{sec:tr}
Trust region methods are a class of iterative optimization
algorithms. In each iteration, an approximate model of the
optimization problem is constructed near the current solution. The
model is only ``trusted'' within a small region around the current
solution called ``trust region'', since in general, approximations fit
the original non-linear function only locally. The approximate model
is then globally optimized within the trust region to obtain a
candidate iterate solution. This step is often called {\em trust
  region sub-problem}.  The size of the trust region is adjusted in
each iteration based on the quality of the current
approximation. Variants of trust region approach differ in the kind of
approximate model used, optimizer for the trust region sub-problem
step and a protocol to adjust the next trust region size. For a
detailed review of trust region methods see \cite{TRreview:Yuan}.

Below we outline a general version of a trust region algorithm in the
context of image segmentation. The goal is to minimize $E(S)$ in
Eq.~\eqref{eq:ERL}. Given solution $S_j$ and distance $d_j$,
the energy $E$ is approximated using
\begin{equation}\label{eq:TR_subproblem}
\widetilde{E}(S) = U_0(S) + \lengthS,
\end{equation}
where $U_0(S)$ is the first order Taylor approximation of the
non-linear term $R(S)$ near $S_j$. The trust region sub-problem is
then solved by minimizing $\widetilde{E}$ within the region given by
$d_j$. Namely, \begin{equation} \label{eq:constrained}
S^*= \underset{||S-S_j||<d}{\operatorname{argmin}} \tilde{E}(S).
\end{equation}
Once a candidate solution $S^*$ is obtained, the quality of the
approximation is measured using the ratio between the actual and
predicted reduction in energy. The trust region is then adjusted accordingly.

For the purpose of our CPU-based evaluations we specifically selected
the Fast Trust Region (FTR) implementation \cite{FTR:cvpr13} which
includes the following components for the trust region framework. The
non-linear term $R(S)$ is approximated by the first order Taylor
approximation $U_0(S)$ in \eqref{eq:TR_subproblem} using first-order
functional derivative \eqref{eq:dR}. The trust region sub-problem in
\eqref{eq:constrained} is formulated as unconstrained Lagrangian
optimization, which is globally optimized using one graph-cut (we use
a floating point precision in the standard code for graph-cuts
\cite{BK:PAMI04}). Note that, in this case, the length term $\lengthS$
is approximated using Cauchy-Crofton formula as in
\cite{GeoCuts:ICCV03}. More details about FTR can be found in
\cite{FTR:cvpr13}.

\subsection{Conceptual Comparison}\label{sec:comp}

\begin{figure*}[t]
\begin{center}
\includegraphics[width = 1\textwidth]{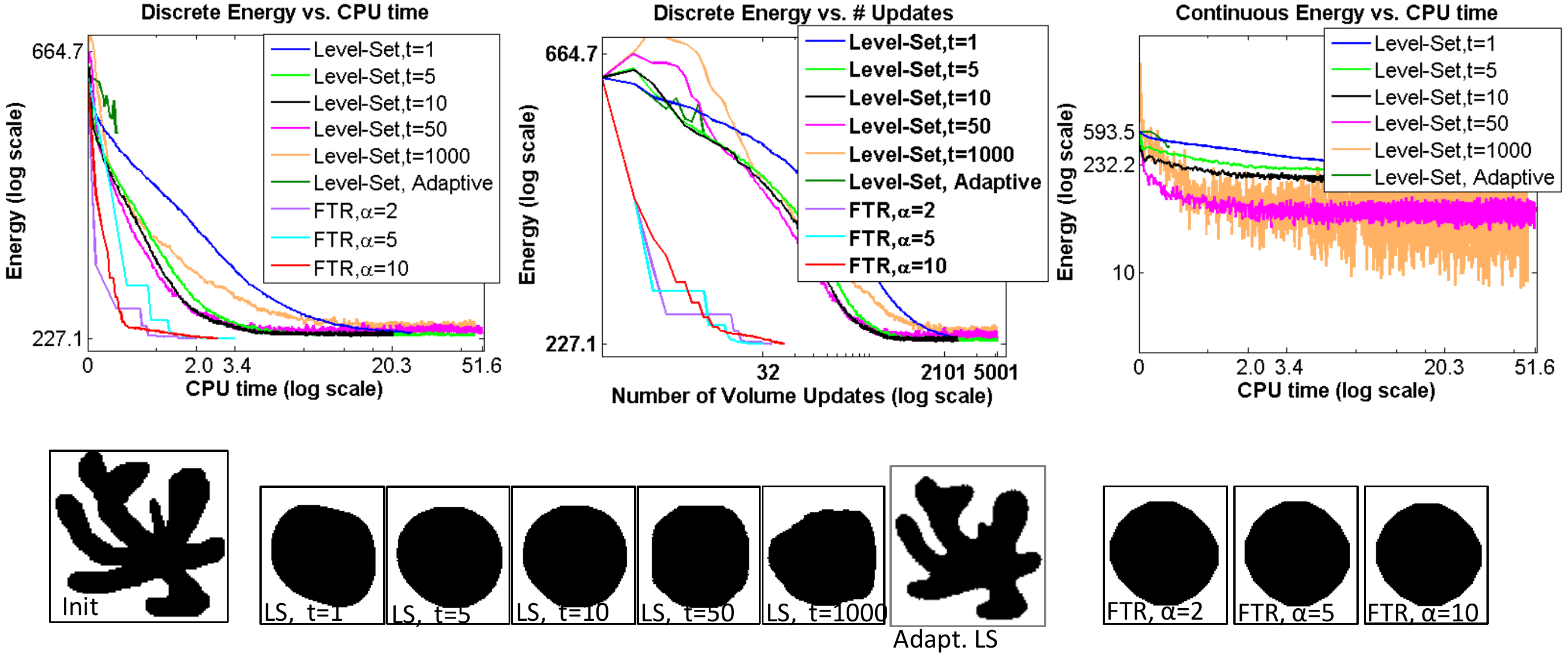} 
\bf\caption{\rm%
  Volume constraint with boundary length regularization. We set the
  weights to $\lambda_{Length}=1$, $\lambda_{Volume}=10^{-4}$.
  \label{fig:volume}}
\end{center}
\end{figure*} 

Some high-level conceptual differences between the {\em level sets}
and {\em trust region} optimization frameworks are summarized in
Figure \ref{fig:yuri_lena}. Standard level sets methods use fixed
$\Delta t$ to make steps $-\Delta t \cdot \frac{\partial E}{\partial
  S}$ in the gradient descent direction. Trust region algorithm
\cite{FTR:cvpr13} moves to solution $S^*$ minimizing approximating
functional $\tilde E(S)$ within a circle of given size $d$. The trust
region size is adaptively changed from iteration to iteration based on
the observed approximation quality. The blue line illustrates the
spectrum of trust region moves for all values of $d$. Solution
$\tilde{S}$ is the global minimum of approximation $\tilde{E}(S)$. For
example, if $\tilde{E}(S)$ is a 2nd-order Taylor approximation of
$E(S)$ at point $S_0$ then $\tilde{S}$ would correspond to a Newton's
step.

\section{Experimental Comparison}\label{sec:experiments}

In this section, we compare trust region and level sets frameworks in
terms of practical efficiency, robustness and optimality.  We selected
several examples of segmentation energies with non-linear regional
constraints. These include: 1) quadratic volume constraint, 2) shape
prior in the form of $L_2$ distance between the target and the observed
shape moments and 3) appearance prior in the form of either $L_2$ distance,
Kullback-Leibler divergence or Bhattacharyya distance between the
target and the observed color distributions.

In all the experiments below, we optimize energy of a general
form $E(S) = R(S) + \lengthS$. To compare optimization quality, we
will plot energy values for the results of both level sets and trust
region. Note that the Fast Trust Region (FTR) implementation uses a
discrete formulation based on graph-cuts and level sets are a
continuous framework. Thus, the direct comparison of their
corresponding energy values should be done carefully.  While numerical
evaluation of the regional term $R(S)$ is equivalent in both methods,
they use completely different numerical approaches to measuring length
$\lengthS$. In particular, level sets use the approximation of length
given in \eqref{eq:LSlength}, while the graph cut variant of trust
region relies on integral geometry and Cauchy-Crofton formula
popularized by \cite{GeoCuts:ICCV03}.

\begin{figure*}[t]
\begin{center}
\includegraphics[width = 0.8\textwidth]{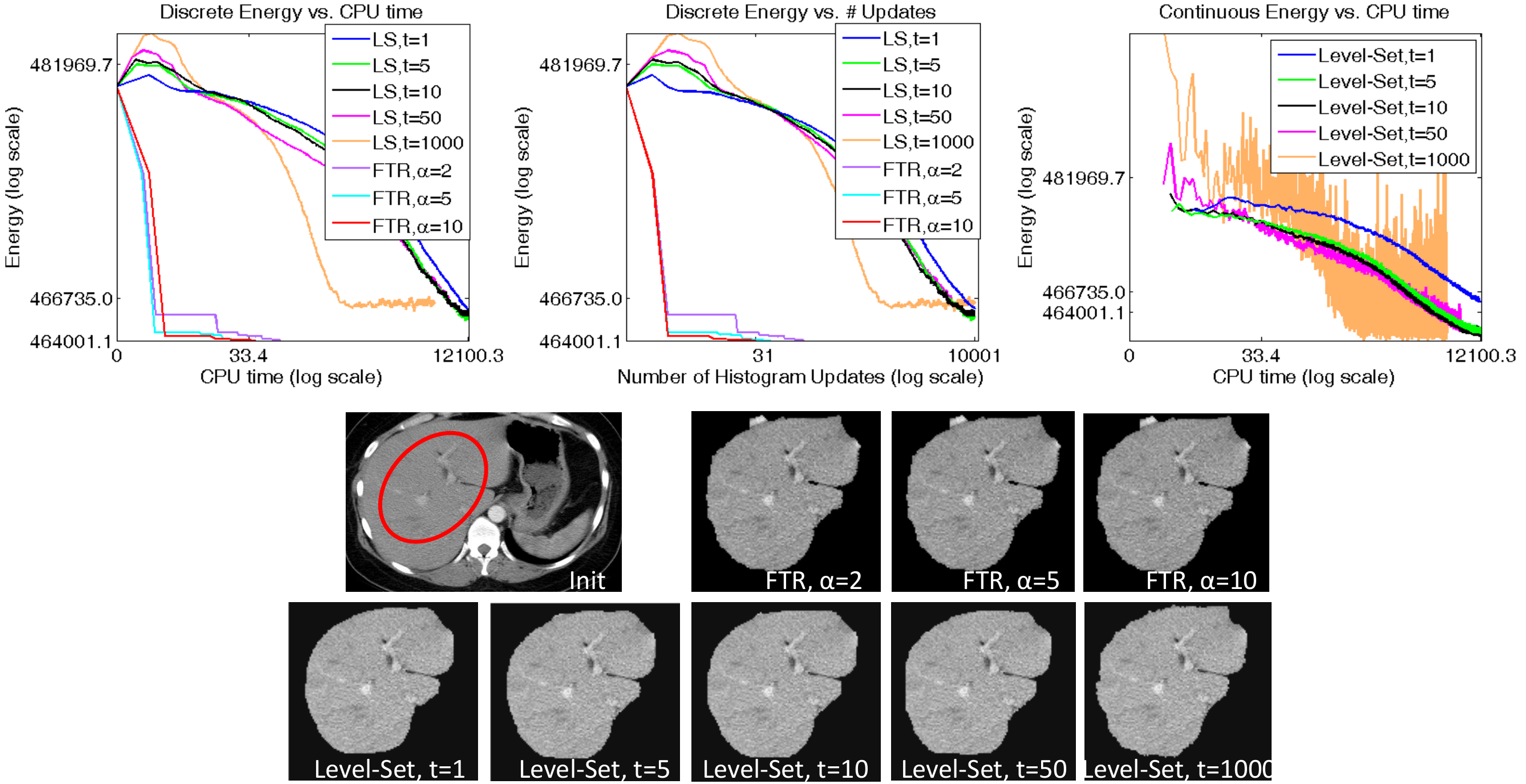} 
\bf\caption{\rm%
  Shape prior constraint with length regularization and log-likelihood
  models. Target shape moments and appearance models are computed from
  the provided ellipse. We used 100 intensity bins, moments up
  to order $l=2$, $\lambda_{Length}=10$, $\lambda_{Shape}=0.01$ and
  $\lambda_{App}=1$. The continuous energy is plotted starting from
  $4^{th}$ iteration to reduce the range of the y-axis.
  \label{fig:liver}}
\end{center}
\end{figure*}

Since the energies are not comparable by its actual number, we study
  instead the robustness of each method independently and compare the
  resulting segmentation with one another. Note that level sets have
  much smaller oscillation for small time steps, which supports the
  theory of the CFL-conditions.

In each application below we examine the robustness of both trust
region and level sets methods by varying the running parameters. In
the trust region method we vary the multiplier $\alpha$ used to change
the size of the trust region from one iteration to another. For the
rest of the parameters we follow the recommendations
of~\cite{FTR:cvpr13}. In our implementation of level sets we vary the
time-step size $\Delta t$, but keep the parameters $\epsilon=1.5$ and
$\mu=0.05$ fixed for all the experiments.

\begin{figure*}[t]
\begin{center}
\includegraphics[width = 0.8\textwidth]{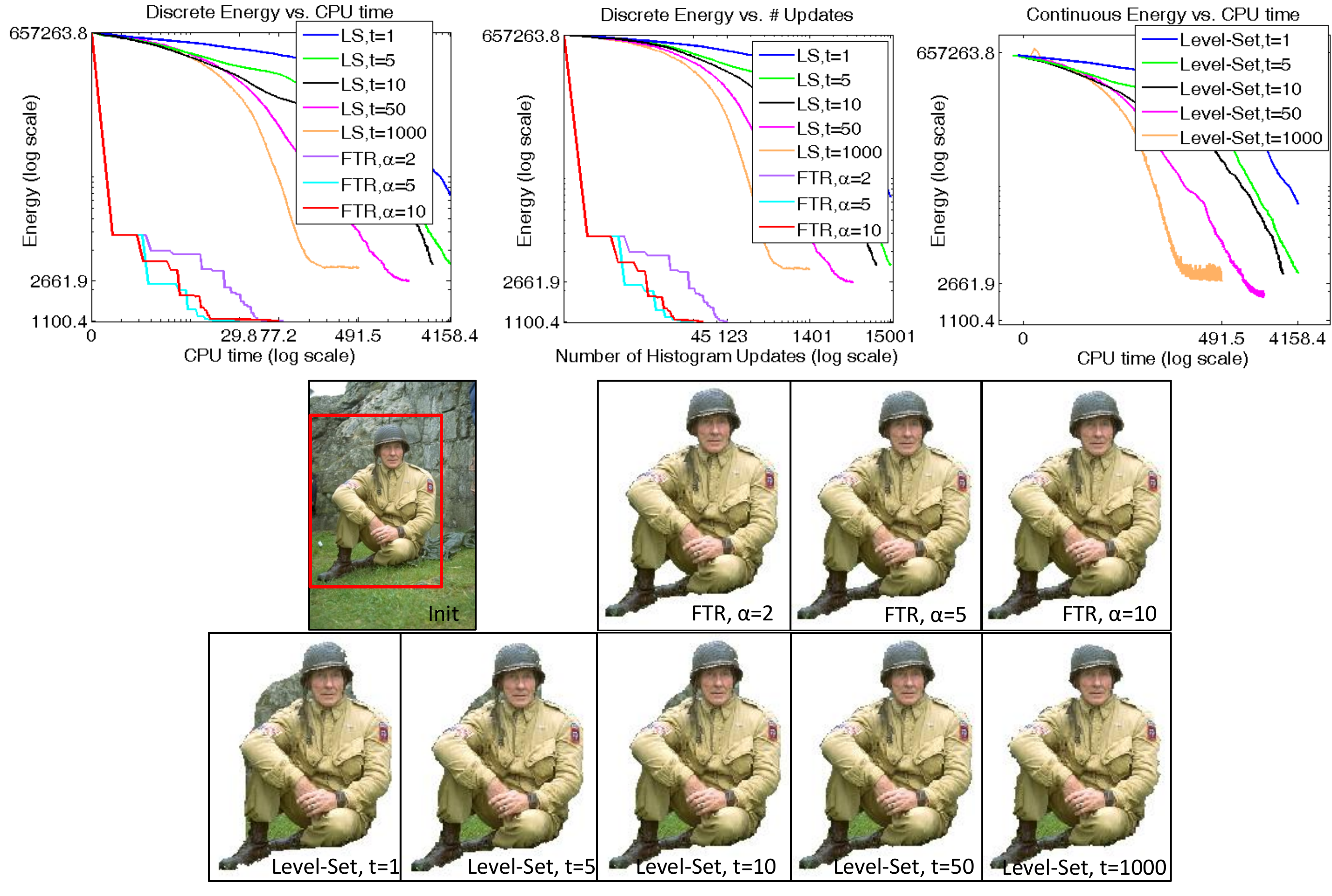} 
\bf\caption{\rm%
  $L_2$ norm between the observed and target color bin counts with
  length regularization. We used 100 bins per channel, $\lambda_{App}
  =1$ and $\lambda_{Length} =1$. \label{fig:soldier_L2L}}
\end{center}
\end{figure*}

The top-left plots in figures \ref{fig:volume}-\ref{fig:mushroom}
report energy $E(S)$ as a function of the CPU time. At the end of each
iteration, both level sets and trust region require energy updates,
which could be computationally expensive. For example,
appearance-based regional functionals require re-evaluation of color
histograms/distributions at each iteration. This is a time consuming
step.  Therefore, for completeness of our comparison, we report in the
top-middle plots of each figure energy $E(S)$ versus the number of
energy evaluations (number of updates) required during the
optimization.
 
\subsection{Volume Constraint}

First, we perform image segmentation with a volume constraint with
respect to a target volume $V_0$, namely,
$$R(S) = (\vecprod{1}{S}-V_0)^2.$$  

We choose to optimize this energy on a synthetic image without
appearance term since the solution to this problem is known to be a
circle.  Figure \ref{fig:volume} shows that both FTR and level sets
converge to good solutions (nearly circle), with FTR being 25 times
faster, requiring 150 times less energy updates and exhibiting more
robustness to the parameters.

\subsection{Shape Prior with Geometric Shape Moments}

Next, we perform image segmentation with a shape prior constraint in the
form of $L_2$ distance between the geometric shape moments of the
segment and a target. Our energy is defined as $E(S) =
\lambda_{Shape}R(S) + \lambda_{Length}\lengthS + \lambda_{App}D(S)$,
where $D(S)$ is a standard log-likelihood unary term based on
intensity histograms. In this case, $R(S)$ is given by
$$
R(S) =\sum_{p+q\leq l}(\vecprod{x^py^q}{S}-m_{pq})^2,
$$
with $m_{pq}$ denoting the target geometric moment of order $l=p+q$.
Figure \ref{fig:liver} shows an example of liver segmentation with the
above shape prior constraint. The target shape moments as well as the
foreground and background appearance models are computed from the user
provided input ellipse as in \cite{KC11:iccv,FTR:cvpr13}. We used
moments of up to order $l=2$ (including the center of mass and shape
covariance but excluding the volume). Both trust region and level sets
obtain visually pleasing solutions. The trust region method is two
orders of magnitude faster and requires two orders of magnitude less
energy updates (top-left and top-middle plots). Since the level sets
method was forced to stop after 10000 iterations, we show the last solution available 
for each value of parameter $\Delta t$ . The
actual convergence for this method would have taken more iterations.
In this example, the oscillations of the energy are especially
pronounced (top-right plot).

\subsection{Appearance Prior}

\exclude{
\begin{figure*}[t]
\begin{center}
\includegraphics[width = 0.8\textwidth]{imgs/soldier.pdf} 
\end{center}
\caption{$L_2$ norm between the observed and target color bin counts, no length regularization ($\lambda_{Length}=0$).  In this case, the discrete and continuous energies are equal. We used 100 bins per channel.\label{fig:soldier_L2}} 
\end{figure*}
}
In the experiments below, we apply both methods to optimize
segmentation energies where the goal is to match a given target
appearance distribution using either the $L_2$ distance between the
observed and target color bin counts, or the Kullback-Leibler
divergence and Bhattacharyya distance between the observed and 
target color distributions.  Here, our energy is defined as $E(S) =
\lambda_{App}R(S) + \lambda_{Length}\lengthS$. We assume $f_i$ is an
indicator function of pixels belonging to bin $i$ and $q_i$ is the
target count (or probability) for bin $i$. The target appearance
distributions for the object and the background were obtained from the
ground truth segments. We used 100 bins per color channel. The images
in the experiments below are taken from \cite{RKB:SIGGRAPH04}.

\paragraph{$L_2$ distance constraint on bin counts:}
Figure \ref{fig:soldier_L2L} shows results of
segmentation with $L_2$ distance constraint between the observed and
target bin counts regularized by length. The regional term in this case is  $$R(S) =
\sqrt{\sum_{i=1}^{k} (\vecprod{f_i}{S}-q_i)^2}.$$ Since the level
sets method was forced to stop after 15000 iterations for values of
$\Delta t=1,5$, we show the last solution available. Full convergence
would have taken more iterations. For higher values of $\Delta t$, we
show results at convergence. We observe two orders of magnitude difference between the
trust region and level sets method in terms of the speed and the
number of energy updates required.

\begin{figure*}[t] 
\begin{center}
\includegraphics[width = 0.8\textwidth]{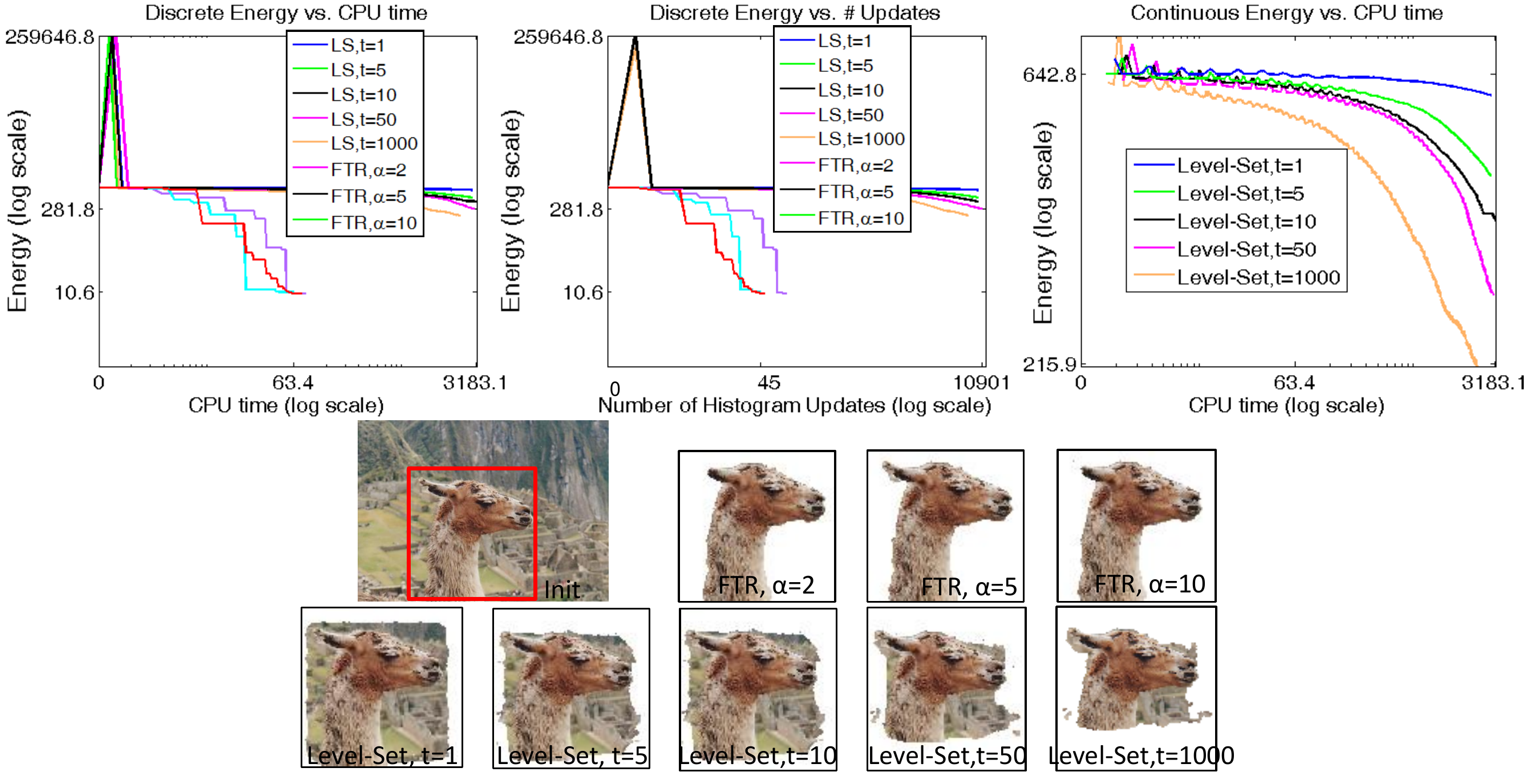} 
\bf\caption{\rm%
  KL divergence between the observed and the target color
  distribution. We used 100 bins per channel, $\lambda_{App}=100$ and
  $\lambda_{Length}=0.01$. Continuous energy is plotted starting from
  forth iteration to reduce the range of the y-axis.
  \label{fig:llama}}
\end{center}
\end{figure*}

\paragraph{Kullback-Leibler divergence:}
Figure \ref{fig:llama} shows results of segmentation with KL
divergence constraint between the observed and target color
distributions. The regional term in this case is given by
$$R(S)=
\sum_{i=1}^k\frac{\vecprod{f_i}{S}}{\vecprod{1}{S}}\log\left(\frac{\vecprod{f_i}{S}}{\vecprod{1}{S}q_i}\right).$$
The level sets method converged for times steps $\Delta t=50$
and $1000$ , but was forced to stop for other values of the parameter
after $10000$ iterations. We show the last solution available. Full
convergence would have required more iterations. The trust region
method obtains solutions that are closer to the ground truth, runs two
order of magnitude faster and requires less energy updates.

\begin{figure*}[t]
\begin{center}
\includegraphics[width = 0.8\textwidth]{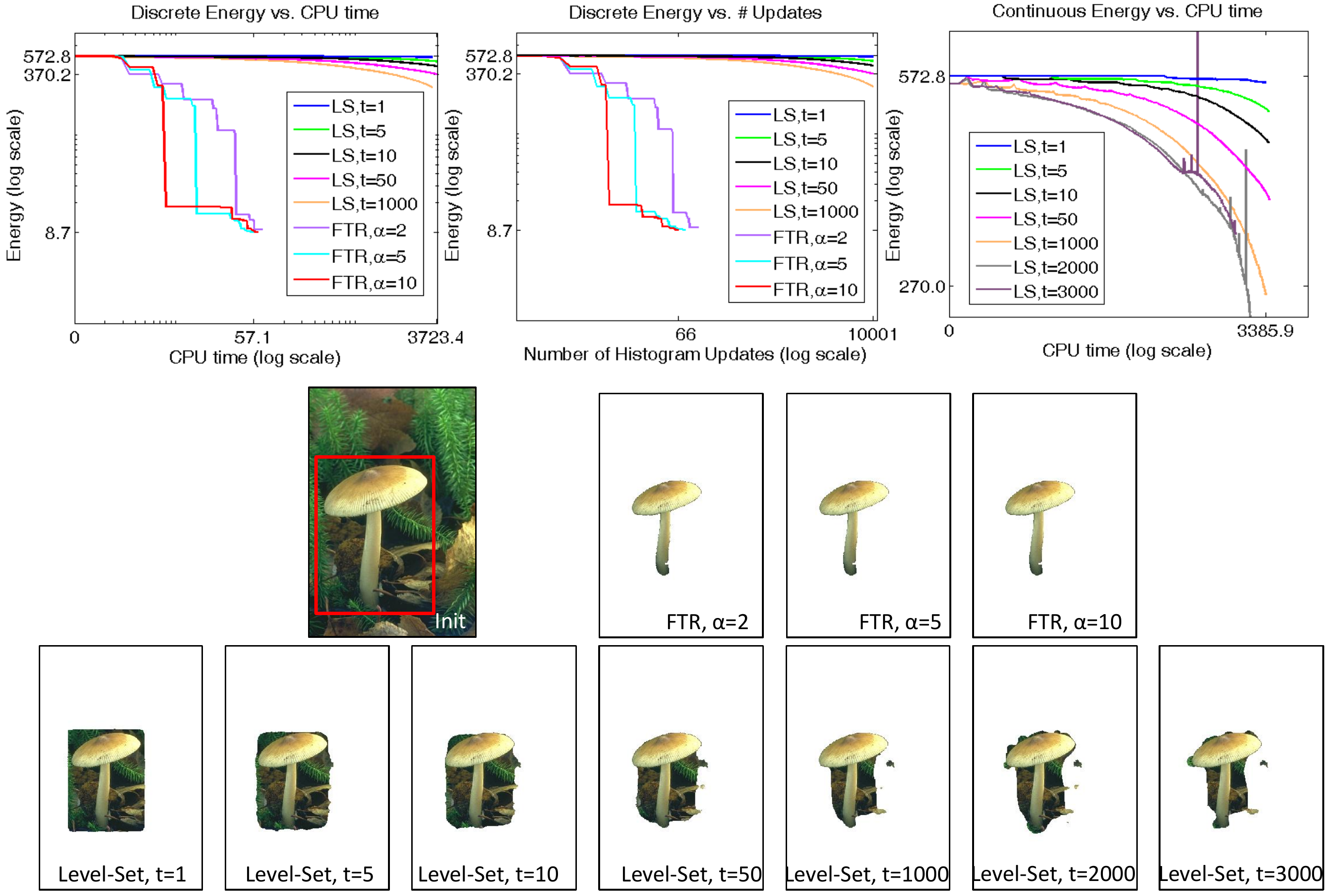} 
\bf\caption{\rm %
  Bhattacharyya distance between the observed and the target color
  distribution. We used 100 bins per channel, $\lambda_{App}=1000$ and
  $\lambda_{Length}=0.01$. \label{fig:mushroom}}
\end{center}
\end{figure*}

\paragraph{Bhattacharyya divergence:}
Figure \ref{fig:mushroom} shows results of segmentation with
Bhattacharyya distance constraint between the observed and target
color distributions. The regional term in this case is given by
$$R(S)= -\log\left(\sum_{i=1}^k
\sqrt{\frac{\vecprod{f_i}{S}}{\vecprod{1}{S}}q_i}\right).$$ Also for
this image, since the level sets method had not (yet) converged after
10000 iterations for any set of the parameters, we show the last
solution available. Further increasing parameter $\Delta t$ would
increase the oscillations of the energy (see top-right plot).

\section{Conclusions}\label{sec:conclusion}
For relatively simple functionals
\eqref{eq:NonlinearRegionalFunctional}, combining a few linear terms
($k$ is small), such as constraints on volume and low-order shape
moments, the quality of the results obtained by both methods is
comparable (visually and energy-wise).  However, we observe that the
number of energy updates required for level sets is two orders of
magnitude larger than for trust region.  This behavior is consistent
with the corresponding CPU running time plots. The segmentation
results on shape moments were fairly robust with respect to parameters
(time step $\Delta t$ and multiplier $\alpha$) for both methods. The
level sets results for volume constraints varied with the choice of
$\Delta t$.  In general, larger steps caused significant oscillations
of the energy in level sets thereby affecting the quality of the
result at convergence.

When optimizing appearance-based regional functionals with large
number of histogram bins (corresponding to large $k$), level sets
proved to be extremely slow.  Convergence would require more than
$10^4$ iterations (longer than 1 hour on our machine). In some cases,
the corresponding results were far from optimal both visually and
energy-wise. This is in contrast to the results by trust region
approach, which consistently converged to plausible solutions with low
energy in less than a minute or $100$ iterations.

\exclude{
\FS{
It is worth noting that the slow convergence of the level set method
is not only due to the curvature flow (see the example without length
in Fig.~\ref{fig:soldier_L2}), but also due to the distance function
penalty in \eqref{distance-function-penalty}. In our experiment we
observed that increasing the time step in level sets does not
necessarily correspond to larger moves within each iteration.
}{}
}

We believe that our results will be useful for many practitioners in
computer vision and medical imaging when selecting an optimization
technique.

{\small
\bibliographystyle{ieee}
\bibliography{arXiv_LevelSetsVsFTR}}

\end{document}